\documentclass[sigconf]{acmart}

\copyrightyear{2024}
\acmYear{2024}
\setcopyright{acmlicensed}
\acmConference[CIKM '24] {Proceedings of the 33rd ACM International Conference on Information and Knowledge Management}{October 21--25, 2024}{Boise, ID, USA.}
\acmBooktitle{Proceedings of the 33rd ACM International Conference on Information and Knowledge Management (CIKM '24), October 21--25, 2024, Boise, ID, USA}
\acmISBN{979-8-4007-0436-9/24/10}
\acmDOI{10.1145/3627673.3680045}

\begin{document}

\title{CPFD: Confidence-aware Privileged Feature Distillation for Short Video Classification}

\author{Jinghao Shi}
\orcid{0009-0003-0196-5111}
\affiliation{%
  \institution{ByteDance}
  \city{San Jose}
  \state{CA}
  \country{USA}
}
\email{jinghao.shi@bytedance.com}

\author{Xiang Shen}
\orcid{0000-0002-6457-0411}
\affiliation{%
  \institution{ByteDance}
  \city{Bellevue}
  \state{WA}
  \country{USA}}
\email{xiang.shen@bytedance.com}

\author{Kaili Zhao}
\orcid{0009-0005-1572-2924}
\affiliation{%
  \institution{ByteDance}
  \city{San Jose}
  \state{CA}
  \country{USA}
}
\email{kaili.zhao@bytedance.com}

\author{Xuedong Wang}
\orcid{0009-0008-0527-8534}
\affiliation{%
  \institution{ByteDance}
  \city{San Jose}
  \state{CA}
  \country{USA}
}
\email{xuedong.wang@bytedance.com}

\author{Vera Wen}
\orcid{0009-0000-7328-409X}
\affiliation{%
  \institution{ByteDance}
  \city{San Jose}
  \state{CA}
  \country{USA}
}
\email{vera.wen@bytedance.com}

\author{Zixuan Wang}
\orcid{0009-0001-5532-5932}
\affiliation{%
  \institution{ByteDance}
  \city{San Jose}
  \state{CA}
  \country{USA}
}
\email{zixuan.wang1@bytedance.com}

\author{Yifan Wu}
\orcid{0009-0008-2798-8860}
\affiliation{%
  \institution{ByteDance}
  \city{San Jose}
  \state{CA}
  \country{USA}
}
\email{yifan.wu@bytedance.com}

\author{Zhixin Zhang}
\orcid{0009-0004-3768-7325}
\affiliation{%
  \institution{ByteDance}
  \city{Beijing}
  \country{China}
}
\email{zhangzhixin.01@bytedance.com}





\renewcommand{\shortauthors}{Jinghao Shi et al.}

\begin{abstract}
Dense features, customized for different business scenarios, are essential in short video classification. However, their complexity, specific adaptation requirements, and high computational costs make them resource-intensive and less accessible during online inference. Consequently, these dense features are categorized as `Privileged Dense Features'.
Meanwhile, end-to-end multi-modal models have shown promising results in numerous computer vision tasks. 
In industrial applications, prioritizing end-to-end multi-modal features, can enhance efficiency but often leads to the loss of valuable information from historical privileged dense features.
To integrate both features while maintaining efficiency and manageable resource costs, we present Confidence-aware Privileged Feature Distillation (CPFD), which empowers features of an end-to-end multi-modal model by adaptively distilling privileged features during training.
Unlike existing privileged feature distillation (PFD) methods, which apply uniform weights to all instances during distillation, potentially causing unstable performance across different business scenarios and a notable performance gap between teacher model (Dense Feature enhanced multimodal-model DF-X-VLM) and student model (multimodal-model only X-VLM), our CPFD leverages confidence scores derived from the teacher model to adaptively mitigate the performance variance with the student model.
We conducted extensive offline experiments on five diverse tasks demonstrating that CPFD improves the video classification F1 score by 6.76\% compared with end-to-end multimodal-model (X-VLM) and by 2.31\% with vanilla PFD on-average. And it reduces the performance gap by 84.6\% and achieves results comparable to teacher model DF-X-VLM. The effectiveness of CPFD is further substantiated by online experiments, and our framework has been deployed in production systems for over a dozen models.
\end{abstract}

\ccsdesc[500]{Computing methodologies~Machine learning}
\ccsdesc[500]{Information systems~Data mining}

\keywords{Short-form Video Classification; Privileged Feature Distillation; Multi-Modal Classification}


\maketitle

\section{Introduction}\label{sec:intro}



Recent years have witnessed the rapid expansion of short video platforms such as TikTok, Youtube Shorts, and Reels. And short video classification is an essential task for short video platforms with a wide range of applications. This task involves tagging videos based on predefined taxonomies, providing insights for user research and being utilized in recommender systems. Additionally, video classification models are instrumental in content moderation, helping to detect inappropriate content. For instance, Qi et al. \cite{qi2023fakesv} curated a dataset from multiple short video platforms focusing on fake news and developed a multi-modal fake news detection model. Das et al. \cite{das2023hatemm} created a multi-modal dataset as a benchmark for hate video classification. Binh et al. \cite{binh2022samba} developed a fusion model to identify inappropriate videos for young children using video titles and metadata.

Video classification predicts video category based on a predefined taxonomy and take a sequence of video frames as input. 
Generally, text information such as title and sticker provides critical information to improve video classification, thus will be integrated to vision features using a multi-modal classification architectures. The performance of the model usually relies on extensive pre-training of foundation models such as CLIP \cite{radford2021learning}, BLIP \cite{li2022blip}, and X-VLM \cite{zeng2022multi} and followed by fine-tuning with task-specific data.

Besides, dense features are also crucial for prediction tasks in industrial applications. These dense features come from different sources. Content-understanding models such as video classification models looking for the existence of specific semantics/objects or text classification models focusing on specific keys words can be one type of source. User interaction features such as likes, shares, and downloads is another type of source. In this work, we enhance the X-VLM model with these dense features to form a Dense Feature-enhanced X-VLM (DF-X-VLM) to empower the model.

However, these dense features are usually inaccessible or prohibitively expensive to access for online services. For example, when predicting the probability of the video containing inappropriate information, the count of user reports only exists after the video accumulates some views, but not immediately available upon video creation. Additionally, maintaining these historical features developed in the previous iterations require significant resources, including memory space for storing the features as well as computational power for inference. Those features are only available during training time but not available during inference and thus we define them as "privileged dense features". On the contrary, features that are available in both training and inference periods such as video frames are defined as non-privileged features. The existence of privileged features could be a severe issue in online machine learning systems at scale.

 Xu and Yang et al. \cite{xu2020privileged,yang2022toward} proposed privileged feature distillation (PFD) to address the issue under the deep neural network (DNN) model framework involving dense privileged features and dense non-privileged features. The PFD approach distill a student model from a teacher model armed with privileged features. However, prior efforts have not integrated PFD into a multi-modal framework. 
 In response, we introduce PFD in multi-modal classification. We employ X-VLM as a student model to learn from DF-X-VLM as a teacher model to harness the information within the dense features without incurring the cost of using those features during online inference.

Conventional PFD methods treat the distillation process uniformly across all instances or adapt solely to the presence of privileged information. However, it is recognized that the performance discrepancies exist across samples due to variations in teacher behavior \cite{zhang2020prime} for general distillation scenario, indicating the distillation performance could be significantly influenced by teacher confidence and thus cause the inherent limitation for conventional previous methods. In the context of distilling multi-modal video classification, we also observe such obvious discrepancies. To be more specific, we found that the performance of student varies a lot in different business tasks especially when the teacher model is empowered with a large number of privileged features. The paper introduces Confidence-aware Privileged Features Distillation for multi-modal video classification, a refined approach to the traditional knowledge distillation process, where the student model not only learns from the output of the teacher model but also leverages additional insights derived from the teacher’s confidence levels. The core idea is to condition the distillation process on the teacher's confidence in its prediction and incorporate uncertainty learning. The shift not only facilitates a deeper understanding and utilization of privileged information but also addresses the performance variances and dependencies associated with traditional distillation and privileged feature approaches.

We summarize our contributions as follows:
\begin{itemize}
    \item We introduce the PFD approach to utilize dense features in a multi-modal classification framework for the first time.
    \item We propose the CPFD algorithm to further address the performance discrepancies introduced by traditional PFD approach.
    \item We validate the effectiveness of the CPFD approach in production data and deploy the model in production. 
\end{itemize}

\section{Related Work}

\subsection{Vision-Language Models}
Video classification has been traditionally viewed as a vision-only task which aims to predict the class of a given input sequence of video frames. Recent work utilized additional available text and audio information to construct a multi-modal classification framework and achieve better performance. The video classification model usually follows pre-training fine-tuning paradigm and benefits from pre-training vision-language foundation models. CLIP 
CLIP \cite{radford2021learning} represents a groundbreaking milestone in vision-language pretraining. , achieving remarkable success in simultaneously understanding images and text via contrastive learning approach. Following this, there has been a surge in the development of Vision-Language Pretraining (VLP) models \cite{dou2022empirical, wang2021simvlm}, which have notably achieved SoTA performance across a range of vision-language benchmarks, such as visual question answering and image captioning assignments.ALBEF \cite{li2021align} incorporates contrastive loss and cross-modal attention to enhance the alignment of image and text representations, facilitating more grounded vision and language representation learning. BLIP \cite{li2022blip} proposed multimodal mixture of encoder-decoder and jointed pretrained by 3 tasks on data precessed by captioning and filtering. X-VLM \cite{zeng2022multi} performed multi-grained vision language pre-training to align the visual concepts in the images and the associated text under multi-granularity.

\subsection{Privileged Features Distillation (PFD)}
A privileged feature is one that is available during model training, but not available at test/online time. Such features naturally arise in merchandised recommendation systems; for instance, “user clicked this item” as a feature is predictive of “user purchased this item” in the offline data, but is clearly not available during online serving. Privileged features distillation (PFD) refers to a natural idea: train a teacher model using all features (including privileged ones) and then use it to train a student model that does not use the privileged features\cite{lopez2015unifying}\cite{yang2022toward}. 
Lopez-Paz et al. \cite{lopez2015unifying} marries the concepts of distillation and privileged information and proposes the initial idea of PFD for the first time. Then Chen et al \cite{xu2020privileged} implemented PFD first time in e-commerce recommendation systems and showcasing how distilling knowledge from privileged features like dwell time can significantly enhance model performance in real-world scenarios. Shuo et al. \cite{yang2022toward}  also explored PFD within the learning-to-rank framework, highlighting its efficacy over traditional methods and analyzing the intricate dynamics between the predictive power of privileged features and student model performance. Similarly, Xiaoqiang et al\cite{gui2024calibration} addresses the challenge of distilling ranking ability for CTR prediction by proposing a calibration-compatible listwise distillation approach. Besides, PFD has also been studied in other areas such as cardinality estimation \cite{page2023node}, noisy-label \cite{ortiz2023does} and entity-relation extraction \cite{zhao2022exploring}
Inspired by all previous works, we also study PFD but in the domain of content moderation and feed quality. We study how privileged features can help content-understand and content-moderation models

\subsection{Confidence-aware}
Although there is no previous methodology which shares the exact same definition of Confidence-aware with this project, there are some similar ideas which inspire our method. In Knowledge Distillation, the confidence of a teacher is used to adaptively assign sample-wise reliability for each teacher prediction with the help of ground-truth labels \cite{zhang2022confidence}. Similarly, the reliability of teacher predictions is used to identify prime samples for distillation and introduce adaptive weighting \cite{zhang2020prime}. In the noisy label domain, Co-teaching designed a two teacher paradigm and used confidence to select the possible clean labels, letting them teach each other\cite{han2018co}. JoCoR also applied a similar approach to utilize the teacher's confidence/loss to adjust the distillation process \cite{wei2020combating}. Besides, work in imitation learning used the idea of confidence-aware, though not specific for the purpose of distillation\cite{zhang2021confidence}. And Meta-Weight-Net designed a weighting function mapping from training loss to sample weight, which is similar to CPFD where we map teacher loss to weight \cite{shu2019meta}.

\section{Methodology}
In this section, we first provide an overview of common short video classification methods, then explain the application of Privileged Feature Distillation (PFD) for short video classification, and introduce our proposed Confidence-aware Privileged Feature Distillation (CPFD), showing its advantages over conventional PFD.
\subsection{Video Classification Overview}
Short video classification requires multimodal understanding. In addition to vision content, text information such as title, sticker, ocr and even audio text also plays an critical role in comprehensively understanding videos. In the industry, video classification methodologies are generally categorized into two types: End-to-end multimodal model and Fusion models.
End-to-end Multimodal models have been one of the most popular areas in machine learning. The proposal of famous works such as ALBEF, BLIP, X-VLM and so on has greatly boosted the application of Multimodal models in video classification model. In this paper, we employed X-VLM\cite{zeng2022multi} as the model architecture because of its outstanding performance in short video classification.


The other type is fusion models which build separate models for each modality to produce dense features with different focus and granularity, and then train fusion models on these dense features. These dense features may focus on understanding content in each modality or specific characteristics, even including non-content-related features such as "video view" (VV) or "video action rate."
Below, we elaborate on the exploration of efficient learning for dense and multimodal features, leading to our final proposals: confidence-aware privilege features (CPFD).

\subsection{Dense Feature enhanced Multimodal Model (DF-X-VLM)}
\begin{figure}
    \centering
    \includegraphics[width=1\linewidth]{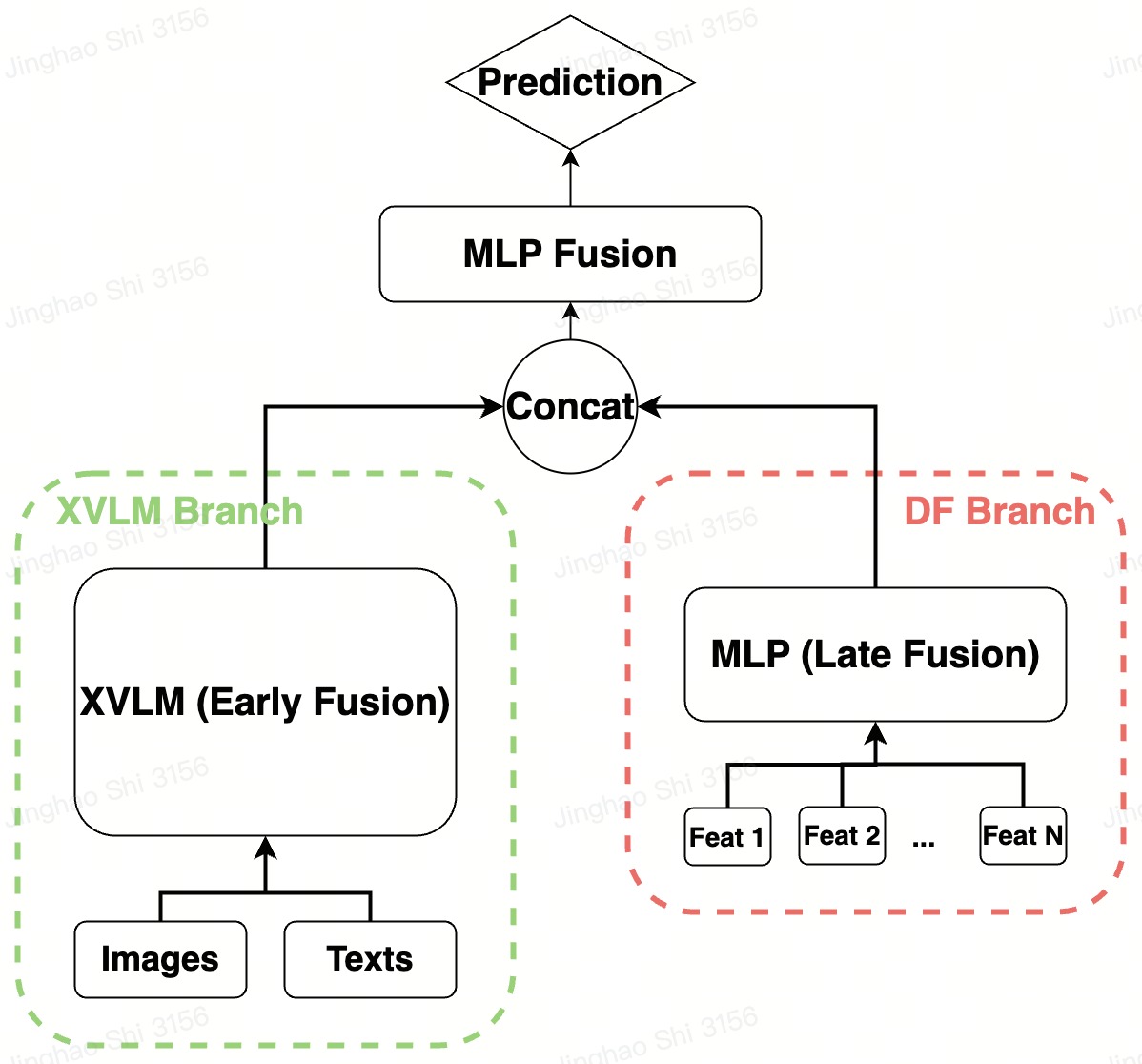}
    \caption{This framework comprises two primary branches. The X-VLM branch represents an end-to-end multi-modal architecture that processes raw images and texts as inputs. The DF branch handles privileged dense features. These two branches are fused together to generate the final prediction.}
    \label{fig1}
\vspace{-10pt}
\end{figure}

It is a natural idea to integrate the multimodal model with the dense feature to use their joint power and make the model address different challenges across multiple content spaces.
We introduce DF-X-VLM. It contains two branchs as illustrated by Fig.~\ref{fig1}: one branch utilizes dense features as inputs (DF-Branch) and the other processes sequence of images and text tokens (X-VLM Branch).The DF branch to designed to memorize the bad contents while the X-VLM branch aims to generalize it. In other words, The DF branch delineate a clear decision boundary of the easy cases and the larger capacity X-VLM branch address the remaining hard cases. 

\subsection{Privileged Feature Distillation (PFD) for Video Classification}
DF-X-VLM approach is an effective method to boost performance However, we still want to use privilege feature distillation (PFD) to train an end-to-end student X-VLM model from DF-X-VLM teacher model for two main reasons:

\begin{figure*}[ht]
    \centering
    \includegraphics[width=0.95\linewidth]{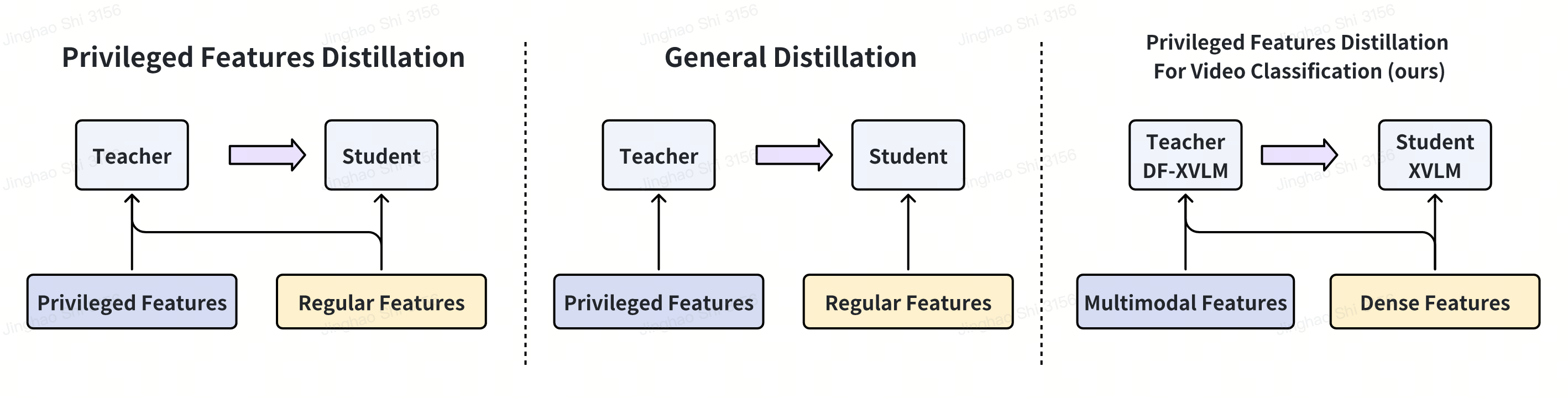}
    \caption{Illustration of General PFD, General Distillation and PFD for Video Classification.}
    \label{fig2}
\end{figure*}

\begin{enumerate}
    \item \textbf{Model Performance}: The X-VLM and DF branches use different coordinate systems and thus the decision boundary is drawn across different content spaces. Unifying the coordinate system may boost performance.
    \item \textbf{Online Pipeline Effciency} Dense features not only lead to more resource consumption, but also result in long chained dependencies which harm online stability. Besides, an end-to-end X-VLM model without any dense feature dependencies makes more rapid scaling across different business scenarios.
\end{enumerate}

\begin{table}[]
    \centering
    \begin{tabular}{lllll}
    \cline{1-4}
        Mean Teacher Loss @ & Neg Sample & Pos Sample & Overall &  \\ \cline{1-4}
        Student Correct     & 0.0057          & 0.8154          & 0.0686  &  \\ 
        Student Incorrect   & 0.2298          & 4.1737          & 0.3806  &  \\ \cline{1-4}
    \end{tabular}
    \caption{Correlation table between Teacher Loss and Student Prediction: This table presents the mean teacher loss, segmented by the correctness of the student's prediction across negative and positive samples.}
    \label{tab:table1}
\vspace{-10pt}
\end{table}
Privileged Features Distillation (PFD) follows a natural distillation strategy: train a “teacher” model using all features (including privileged ones) and then use it to train a “student” model that does not use the privileged features. A comparison between General PFD, General Distillation and PFD for video classification is shown in Fig.~\ref{fig2}. A privileged feature is one that is available during model training, but not available at test/online time. In the context of multimodal video classification, we define all historical dense features that is expensive to use as "privileged features". See more detailed definition in Section \ref{sec:intro}.  For actual implementation, we distill an X-VLM model (student) from a DF-X-VLM model (teacher): 
\begin{itemize}
    \item The student model will be trained using a paramatrized combination of classification loss and distillation loss, which is a common practice in knowledge distillation: 
    \begin{equation}
        \mathcal{L}_{student}=(1-\alpha)*\mathcal{L}_{cls}+\alpha*\mathcal{L}_{distill}
    \end{equation}
    \begin{equation}
        \mathcal{L}_{cls}=H(q_s, y) = -\sum_{i} y_{i} \log(q_{s,i})
    \end{equation}
    \begin{equation}
        \mathcal{L}_{distill}=D_{\text{KL}}(q_t || q_s) = T^2 \sum_{i} q_{t,i}(T) \log\left(\frac{q_{t,i}(T)}{q_{s,i}(T)}\right)
    \end{equation}
    \item Where:
    \begin{itemize}
        \item $\alpha$ represents the scaling factor balancing two losses
        \item $T$ is the temperature parameter
        \item $q_{s}$ is the predicted probability distribution generated by the student.
        \item $q_{t}$ is the soft target probability distribution generated by the teacher.
        \item $y$ is the hard label.
    \end{itemize}
\end{itemize}

Note that though we adopt X-VLM and DF-X-VLM as our use case, the PFD framework is suitable for any model architecture. With that said, the teacher and student can be any model architecture and are totally independent of one another.

\subsection{Confidence-aware Privileged Feature Distillation (CPFD)}
\begin{figure}
    \centering
    \includegraphics[width=1\linewidth]{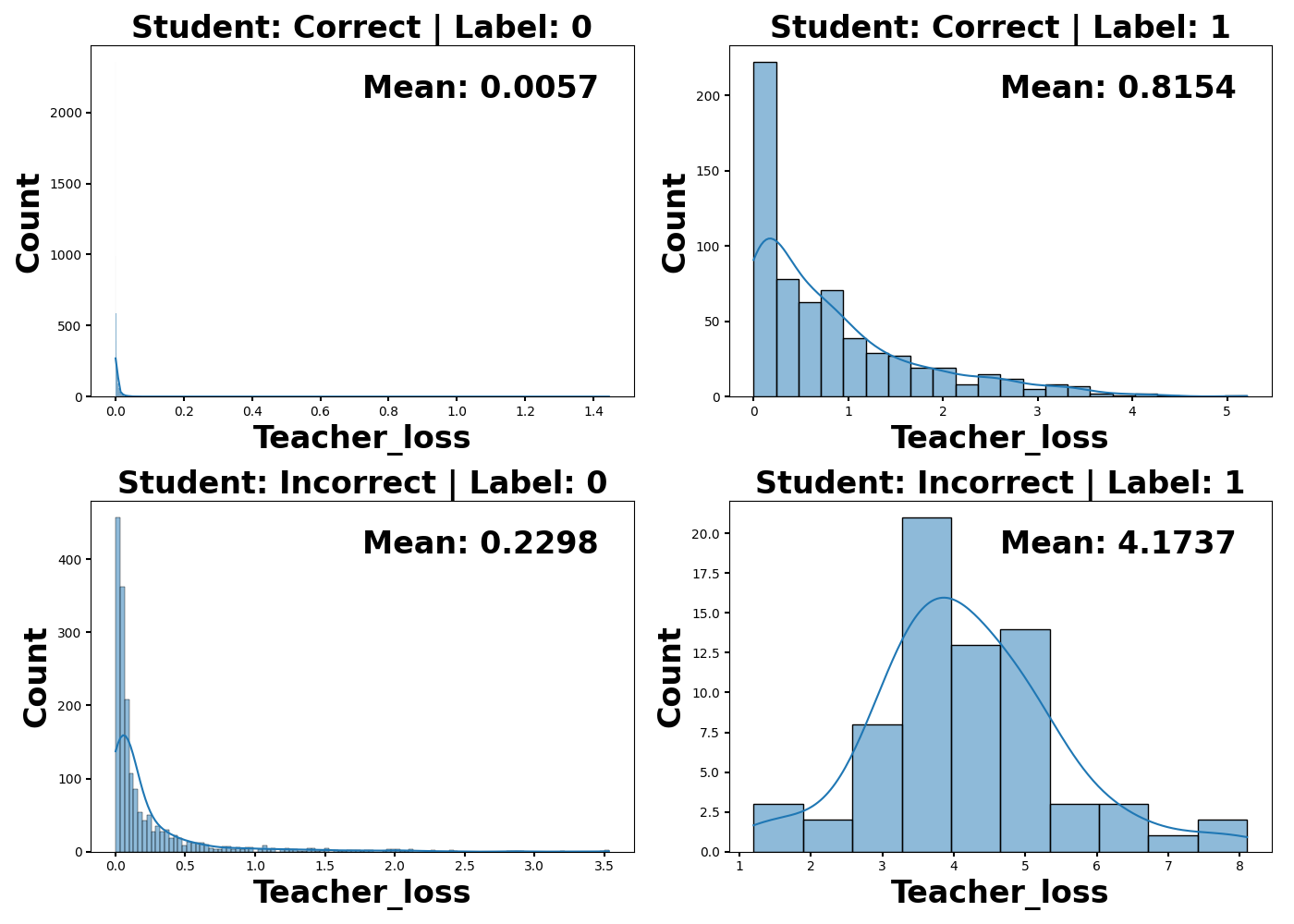}
    \caption{Correlation plot between Teacher Loss and Student Prediction: This table presents the teacher loss distribution, segmented by the correctness of the student's prediction across negative and positive samples.}
    \label{fig3}
\end{figure}

Though PFD proves to be an effective method, there are still performance gaps between teacher and student, especially when the teacher model becomes more complicated. Below we analyze the causes of those performance gaps and propose a refined approach to address the gaps.
\begin{itemize}
    \item In traditional distillation framework, the student learning is often uniformly influenced by the teacher’s outputs across all examples. It does not take into account the varying levels of certainty or confidence the teacher possesses in its own predictions. 
    \item To analyze how the confidence of the teacher influences the student's behavior, we quantified the teacher's confidence through the distribution of teacher loss associated with different student predictions (Fig.~\ref{fig3} and Table~\ref{tab:table1}). They highlights the varying impact of prediction accuracy on teacher loss, illustrating significant differences between correct and incorrect predictions by the student. We can conclude that the student prone to make mistakes when the teacher loss is high. And it is extremely obvious for false negative cases.
\end{itemize}

\begin{figure*}[ht]
    \centering
    \includegraphics[width=0.85\linewidth]{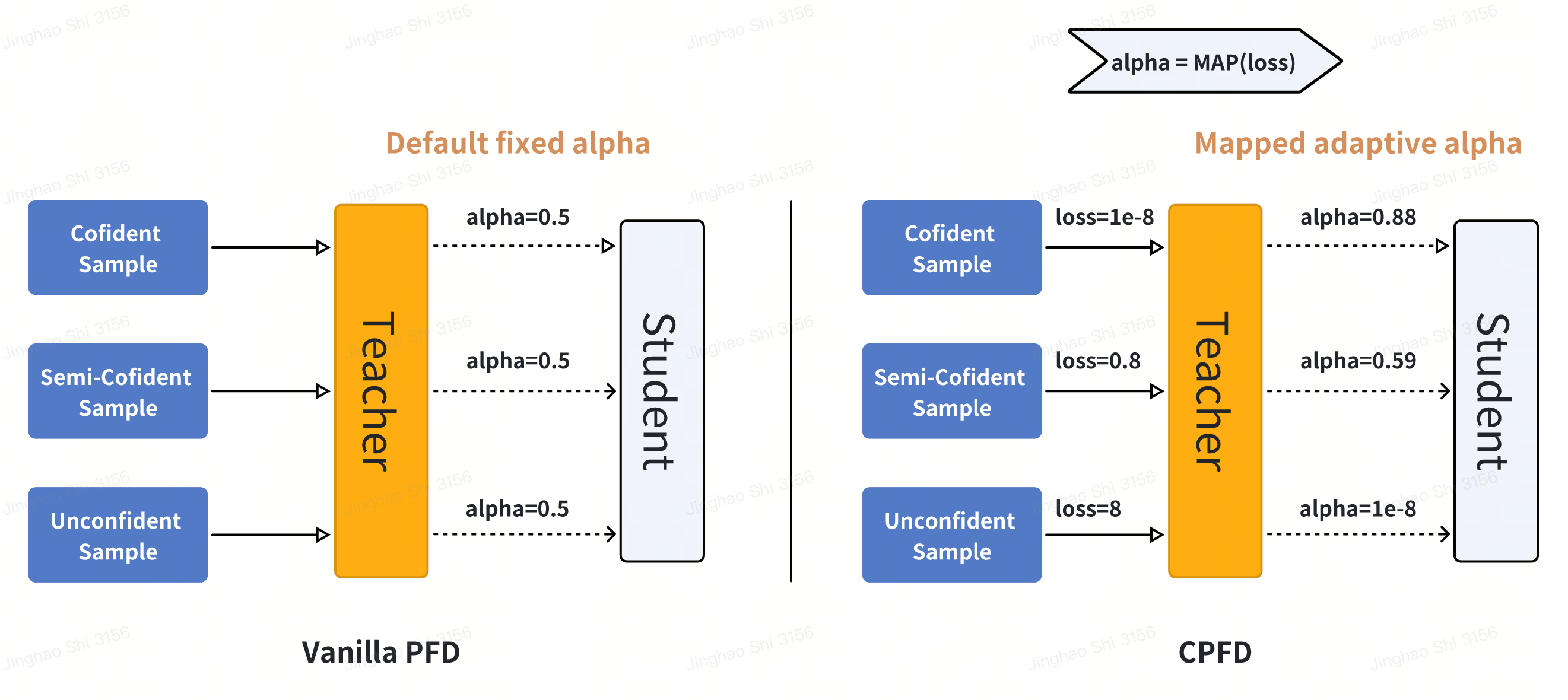}
    \caption{Comparison between CPFD and Vanilla PFD. Vanilla PFD applies a uniform alpha value across all samples. In contrast, CPFD adaptively adjusts the alpha parameter based on the teacher's loss value for each sample.}
    \label{fig4}
\end{figure*}

\subsubsection{Redesign PFD to Incorporate Teacher Confidence} 

Based on the above analysis, we propose Confidence-aware Privileged Feature Distillation (CPFD) as shown in Fig~\ref{fig4}. CPFD is a refined approach to the traditional knowledge distillation process, where the student model not only learns from the output of the teacher model but also leverages additional insights derived from the teacher’s confidence levels. The core idea behind this methodology is to condition the distillation process on how certain the teacher is about its predictions and incorporating uncertainty learning. \\
The proposed confidence-aware mechanism aims to tailor the student’s learning experience by weighting the distillation loss according to the teacher’s confidence, thereby mimicking the teacher more closely when it is more certain, and relying more on the ground truth when the teacher’s confidence is low.
This approach stems from the observation that not all knowledge from the teacher is equally reliable. By modulating the distillation process based on teacher confidence, the student can potentially avoid learning from the teacher's less reliable cues, which may lead to a more robust and accurate student model.
The standard loss function for vanilla PFD is as below:
    \begin{equation}
        \mathcal{L}_{student}=(1-\alpha)*\mathcal{L}_{cls}+\alpha*\mathcal{L}_{distill}
    \end{equation}
The combination of classification loss and distillation loss is a common practice in knowledge distillation. In the context of video understanding, a new angle to interpret the loss combination is to take it as a multi-teacher distillation. 
\begin{itemize}
    \item The "distillation teacher" who is easier to learn from but also tends to have information loss. Moreover, the teacher is not same-level confident in all cases. To be more specific, the teacher can be good at some aspects while bad at others
    \item The "label teacher" who is hard to learn from but has no information loss. However, this teacher would also suffer from noise-label problems, which is quite common in scenarios of video understanding.
\end{itemize}
And this raises the natural questions to traditional PFD: To what extent we should trust each teacher? And the answer is we should adaptively trust them. Based on this, we adjusted the standard loss function to dynamically control the $\alpha$ who balance the distillation loss and classification loss. For actual implementation of CPFD, we used teacher's loss to represent confidence, which is used to control the $\alpha$, representing how much we would learn from the teacher on this sample. A comparison between CPFD and PFD is shown in Fig.~\ref{fig4}

\subsubsection{Implementing CPFD}
\begin{figure}
    \centering
    \includegraphics[width=1\linewidth]{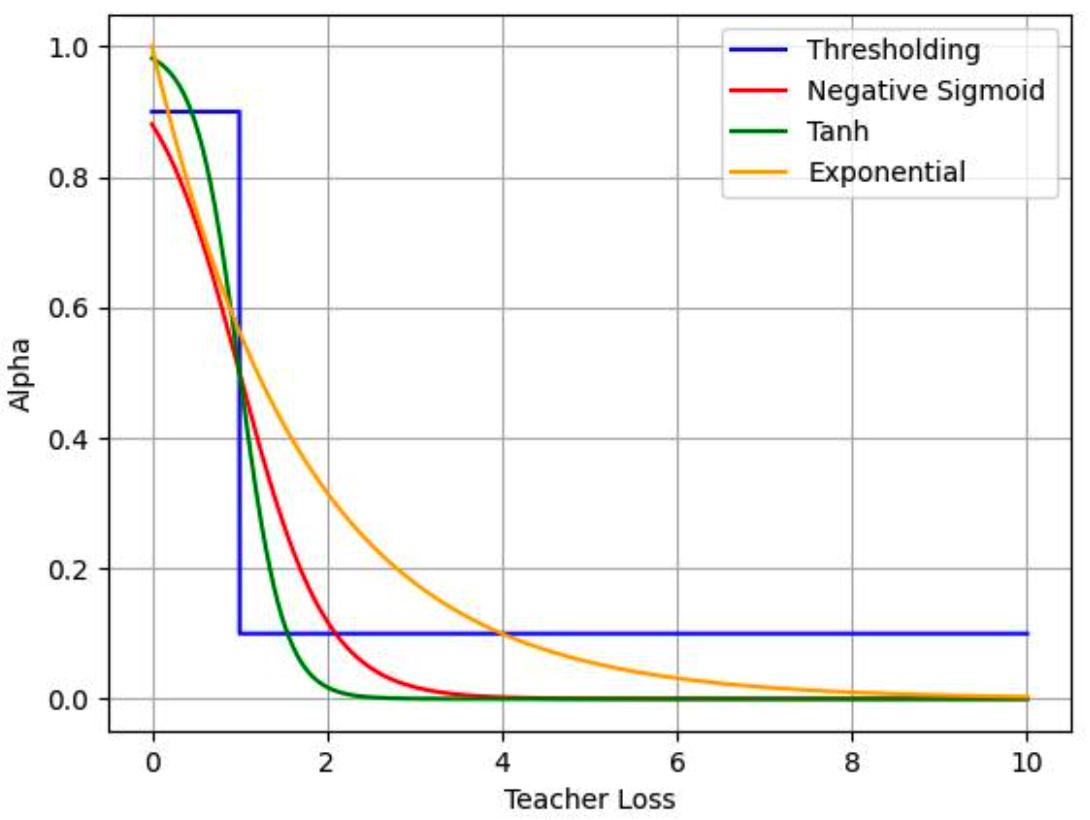}
    \caption{Alpha Distribution as a Function of Teacher Loss.}
    \label{fig5}
\vspace{-10pt}
\end{figure}

To operationalize this concept, we designed several mapping mechanisms to allow the loss to control the $\alpha$. The detailed mapping function and distribution are outlined in Table~\ref{tab:table2} and its distribution are shown Fig.~\ref{fig5}.

\begin{table}[h]
\centering
\begin{tabular}{|c|c|}
\hline
\textbf{Mapping} & \textbf{Description} \\ \hline
Threshold & $\alpha = 0.9 \text{ if } L_{\text{teacher}} < \tau \text{ else } 0.1$ \\ \hline
Neg Sigmoid & $\alpha = \frac{1}{1 + e^{\beta \cdot (L_{\text{teacher}} - L_{\text{center}})}}$ \\ \hline
Tanh & $\alpha = 0.5 \cdot (\tanh(-\beta \cdot (L_{\text{teacher}} - L_{\text{center}})) + 1)$ \\ \hline
Exp Decay & $\alpha = \alpha_{\max} \cdot e^{-k \cdot (L_{\text{teacher}} - L_{\text{min}})}$ \\
& where $k = -\log\left(\frac{\alpha_{\min}}{\alpha_{\max}}\right) / (L_{\max} - L_{\text{min}})$ \\ \hline
\end{tabular}
\caption{Mapping Function Designs.}
\label{tab:table2}
\vspace{-20pt}
\end{table}

\section{Experiments}
In this section, we conduct experiments to evaluate both the offline and online A/B performance of the video classification models trained by PFD and CPFD.

\begin{table*}[ht]
\centering
\begin{tabular}{|c|ccc|ccc|ccc|ccc|}
\hline
\textbf{Task} & \multicolumn{3}{c|}{\textbf{X-VLM}} & \multicolumn{3}{c|}{\textbf{DF-X-VLM (Teacher)}} & \multicolumn{3}{c|}{\textbf{PFD-XVLM (Student)}} & \multicolumn{3}{c|}{\textbf{CPFD-X-VLM (Student)}} \\
& ROC & PR & F1 & ROC & PR & F1 & ROC & PR & F1 & ROC & PR & F1 \\
\hline
Task1 & 0.9269 & 0.5621 & 0.5424 & 0.9493 & 0.6878 & 0.6277 & 0.9400 & 0.6584 & 0.6104 & \textbf{0.9460} & \textbf{0.6667} & \textbf{0.6266} \\
Task2 & 0.9515 & 0.9646 & 0.8073 & 0.9616 & 0.9720 & 0.8223 & 0.9529 & 0.9657 & 0.8191 & \textbf{0.9555} & \textbf{0.9670} & \textbf{0.8303} \\
Task3 & 0.9927 & 0.9739 & 0.9255 & 0.9962 & 0.9844 & 0.9443 & 0.9928 & 0.9744 & 0.9315 & \textbf{0.9935} & \textbf{0.9764} & \textbf{0.9337} \\
Task4 & 0.9657 & 0.9541 & 0.7767 & 0.9728 & 0.9621 & 0.8009 & 0.9697 & 0.9587 & \textbf{0.7964} & \textbf{0.9716} & \textbf{0.9614} & 0.7962 \\
Task5 & 0.9439 & 0.4100 & 0.4538 & 0.9556 & 0.4515 & 0.4942 & 0.9489 & 0.4418 & 0.4737 & \textbf{0.9550} & \textbf{0.4728} & \textbf{0.5083} \\
\hline
\end{tabular}
\caption{Performance Metrics Across Different Models (ROC:ROC AUC, PR:PR AUC, F1:F1 Score). X-VLM refers to normally trained X-VLM model. PFD-X-VLM and CPFD-X-VLM refers to student X-VLM model distilled from teacher DF-X-VLM using vanilla PFD method and Confidence-aware PFD method. The bold number represents the best \underline{student} model results.}
\label{tab:table3}
\end{table*}

\subsection{Experiment Setting}
\textbf{Data.} Since the proposed CPFD mainly focus on the industrial video classification models, we conduct the experiments on real-world industrial datasets. The dataset spans five diverse classification tasks in different domains. For each classification task, we collected 5 to 20 millions of videos in the past 3 months. The labels under different taxonomies are human labeled. We extract 5\%-10\% data as the evaluation dataset and leave the rest as training dataset based on the available amount of data of each task. \\
\textbf{Model.} As mentioned, we employ X-VLM architecture \cite{zeng2022multi} as our primary model for video classification. The model takes two type of inputs: visual and text. For visual input, we extract and concatenate video frames. For text input, we concatenate the video title with the overlay text within the video. The implementation is modified based on the code base released by X-VLM (https://github.com/zengyan-97/X-VLM). \\
\textbf{Training.} Training commences from a in-house checkpoint pre-trained on over 5 billion short video data. We start with an initial learning rate of 1e-5, which decays over time. Models are trained on a single 8-A100 node, with training time ranging from 1 to 5 days depending on the data volume. \\
\textbf{Baselines.} To verify the power of the proposed CPFD, Our proposed CPFD is benchmarked against the following models:
\begin{itemize}
    \item \textbf{X-VLM}. Normally trained X-VLM model without any distillation. It is basically a reproduction of the paper X-VLM with some minor modification to adapt to business need.
    \item \textbf{DF-X-VLM}. Dense Feature enhance X-VLM as described in Section 3.2 and Fig.~\ref{fig1}.
    \item \textbf{PFD-X-VLM}. A X-VLM student model distilled from teacher DF-X-VLM using vanilla PFD method.
\end{itemize}

To ensure a fair comparison, all methods are implemented on same code base, model starting checkpoint, learning rate, optimizer and other hyper-parameters based on empirical observations. 

\subsection{Offline Evaluation Results}
\textbf{Offline Evaluation Metrics.} To show a comprehensive comparison between proposed method and baselines, we compare ROC AUC, PR AUC and F1 score. PR AUC is particularly informative for evaluating binary classifiers on imbalanced datasets\cite{saito2015precision} \\

\begin{table}[]
\begin{tabular}{llllll}
\hline
Task                          & Task1 & Task2 & Task3 & Task4 & Task5 \\ \hline
Cost Ratio & 4.08  & 3.5   & 9     & 4.5   & 3.92  \\ \hline
\end{tabular}
\caption{Additional Dense Feature Inference Cost compared with X-VLM model.}
\label{tab:table4}
\vspace{-20pt}
\end{table}

\subsubsection{Performance Comparison}
Results are detailed in Table~\ref{tab:table3}, from which we draw two main conclusions:
\begin{itemize}
    \item Proposed CPFD consistently outperforms other methods across all metrics and all tasks. Specifically, CPFD improves video classification F1 score by 6.76\% over the standard X-VLM and by 2.31\% over PFD. 
    \item Relatively, CPFD significantly reduced the performance F1 score gap between X-VLM and DF-X-VLM by 84.6\% and that between PFD-X-VLM and DF-X-VLM by 62.2\%. CPFD performs very closely with the teacher model DF-X-VLM, indicating minimum information loss removing the DF branch through distillation.
\end{itemize}

\begin{table}[]
\begin{tabular}{lllll}
\hline
Model               & X-VLM   & DF-X-VLM & PFD  & CPFD    \\ \hline
Hit Rate            & 19.82\% & 28.00\%  & 26.88\%  & 28.00\%  \\
Relative Difference & -       & +41.27\% & +35.62\% & +41.27\% \\
\hline
\end{tabular}
\caption{Performance Metrics Across Different Models in Online A/B experiment. PFD represents PFD X-VLM and CPFD represents CPFD X-VLM.}
\label{tab:table5}
\vspace{-10pt}
\end{table}

\subsubsection{Ablation Study} In this section, we study the influence of different mapping function designs and temperature setting during distillation \\
\textbf{Mapping Functions.} We evaluated four empirically derived mapping functions, as shown in Table~\ref{tab:table2}. The four mapping functions including the thresholds and parameters we provided here are purely empirical, inspired by some popular activation function based on some basic facts:
\begin{itemize}
    \item The output range needs to between 0 and 1
    \item The function should be a monotonically increasing function with the loss input.
\end{itemize}

We experimented with all four mapping functions on Task1,2 and 3. he following observations can be drawn based on results in Table~\ref{tab:table6} 
\begin{itemize}
    \item There is no definitive best performer, though Exponential Decay functions yields slightly better outcomes due to varying data distributions.
    \item The performance of different mapping design is relatively stable and consistently outperforms X-VLM and PFD-X-VLM
    \item Mapping functions and their parameters should be selected according to the specific data distribution and problem context. While we present several promising options here, there may be more optimal mapping function designs that could further enhance performance
\end{itemize}

\begin{table}[ht]
\centering
\begin{tabular}{lcccc}
\hline
 & \textbf{Threshold} & \textbf{Neg Sigmoid} & \textbf{Tanh} & \textbf{Exp Decay} \\
\hline
Task1 & 0.9453 & 0.9441 & 0.9439 & 0.9460 \\
Task2 & 0.9933 & 0.9932 & 0.9933 & 0.9935 \\
Task3 & 0.9550 & 0.9549 & 0.9550 & 0.9542 \\
\hline
\end{tabular}
\caption{Comparison of Model AUC with Different Mapping Functions across Tasks.}
\label{tab:table6}
\vspace{-10pt}
\end{table}

\textbf{Temperature.} We also ablated the important parameter Temperature during distillation \cite{hinton2015distilling}. Due to space limitations, we only display the experimental results for Task1. The results in Table~\ref{tab:table7}. show that T=1 works the best in our setting. So to make a fair comparison, we used 1 as the default temperature across all our experiments. 

\begin{table}[ht]
\centering
\begin{tabular}{lccc}
\hline
 & \textbf{T=0.5} & \textbf{T=1} & \textbf{T=2} \\
\hline
ROC AUC & 0.9402 & 0.9460 & 0.9439 \\
\hline
\end{tabular}
\caption{Comparison of Model AUC with Different Temperature on Task1.}
\label{tab:table7}
\vspace{-10pt}
\end{table}

\subsubsection{Computational Cost}
We calculate the required online resources to inference the dense features compared with the inference cost of X-VLM models among all 5 different tasks and present the results in Table~\ref{tab:table4}. On average, the DF-X-VLM model spends 5 times more resources compared with the X-VLM model among the 5 tasks while some of the features are counted repeated when used in different tasks. It shows that CPFD is a better option due to comparable classification performance and lower computation resources. 

In addition to resource consumption, CPFD also significantly improve the online service stability. It liberates the model from the dependency on specific features during online inference, allowing reliance solely on raw inputs such as video, text, or audio and mitigating online-offline model performance difference. Further more, it largely enhanced the model capability by granting the flexibility to leverage a broader range of privileged features during training, without the complications associated with online service and deployment.  Lastly, it reduces operational costs and simplifies the expansion of applications.

\subsection{Online Experiments}
We conduct an online A/B experiment to further evaluate the effectiveness of the proposed approach. Specifically, we deployed 4 different models targeting Task 3 in the production system and the models are evaluated in Table~\ref{tab:table5}. We employ the X-VLM model as the control group and conduct the experiment for 5 consecutive days. We set specific model score thresholds for each model and created 4 different model based rules. New videos published with model scores higher than the thresholds will be recalled. Human reviewers will review all the recalled videos and decide if the videos are accurate. We adjust the thresholds to minimize the differences of total videos recalled per day between the baseline model and other models. During the experiment, we evaluate the results by the hit rate as the number of positive videos after human review within recalled videos by the number recalled videos. The result is presented in Table~\ref{tab:table5}.

From the result, we could observe PFD X-VLM significantly improves performance compared with X-VLM. CPFD XVLM further improves the performance and achieve same hit rate as the teacher model DF-X-VLM, again indicating almost no information loss during distillation 

\section{Conclusion}
In this paper, we introduce a Confidence-aware Privileged Feature Distillation (CPFD) approach tailored for video classification. This method effectively utilizes information from privileged dense features without incurring additional inference costs. Experimented on real-world datasets, both offline and online evaluation results consolidate the effectiveness of CPFD. This framework not only enhances classification performance but also maintains operational efficiency. It has been successfully integrated into production systems with multiple deployed models.

\begin{acks}
We extend our gratitude to Zeya Wang for his invaluable assistance during the initial stages of this project. We also thank Zhiqian Chen and Kenan Xiao for their contributions to the offline experiments. Special thanks go to Ardalan Mehrani, Zhongrong Zuo, and Chengkai Jin for their insightful technical discussions.
\end{acks}

\bibliographystyle{ACM-Reference-Format}
\balance
\bibliography{reference}


\begin{thebibliography}{24}


\ifx \showCODEN    \undefined \def \showCODEN     #1{\unskip}     \fi
\ifx \showDOI      \undefined \def \showDOI       #1{#1}\fi
\ifx \showISBNx    \undefined \def \showISBNx     #1{\unskip}     \fi
\ifx \showISBNxiii \undefined \def \showISBNxiii  #1{\unskip}     \fi
\ifx \showISSN     \undefined \def \showISSN      #1{\unskip}     \fi
\ifx \showLCCN     \undefined \def \showLCCN      #1{\unskip}     \fi
\ifx \shownote     \undefined \def \shownote      #1{#1}          \fi
\ifx \showarticletitle \undefined \def \showarticletitle #1{#1}   \fi
\ifx \showURL      \undefined \def \showURL       {\relax}        \fi
\providecommand\bibfield[2]{#2}
\providecommand\bibinfo[2]{#2}
\providecommand\natexlab[1]{#1}
\providecommand\showeprint[2][]{arXiv:#2}

\bibitem[Binh et~al\mbox{.}(2022)]%
        {binh2022samba}
\bibfield{author}{\bibinfo{person}{Le Binh}, \bibinfo{person}{Rajat Tandon}, \bibinfo{person}{Chingis Oinar}, \bibinfo{person}{Jeffrey Liu}, \bibinfo{person}{Uma Durairaj}, \bibinfo{person}{Jiani Guo}, \bibinfo{person}{Spencer Zahabizadeh}, \bibinfo{person}{Sanjana Ilango}, \bibinfo{person}{Jeremy Tang}, \bibinfo{person}{Fred Morstatter}, {et~al\mbox{.}}} \bibinfo{year}{2022}\natexlab{}.
\newblock \showarticletitle{Samba: Identifying Inappropriate Videos for Young Children on YouTube}. In \bibinfo{booktitle}{\emph{Proceedings of the 31st ACM International Conference on Information \& Knowledge Management}}. \bibinfo{pages}{88--97}.
\newblock


\bibitem[Das et~al\mbox{.}(2023)]%
        {das2023hatemm}
\bibfield{author}{\bibinfo{person}{Mithun Das}, \bibinfo{person}{Rohit Raj}, \bibinfo{person}{Punyajoy Saha}, \bibinfo{person}{Binny Mathew}, \bibinfo{person}{Manish Gupta}, {and} \bibinfo{person}{Animesh Mukherjee}.} \bibinfo{year}{2023}\natexlab{}.
\newblock \showarticletitle{Hatemm: A multi-modal dataset for hate video classification}. In \bibinfo{booktitle}{\emph{Proceedings of the International AAAI Conference on Web and Social Media}}, Vol.~\bibinfo{volume}{17}. \bibinfo{pages}{1014--1023}.
\newblock


\bibitem[Dou et~al\mbox{.}(2022)]%
        {dou2022empirical}
\bibfield{author}{\bibinfo{person}{Zi-Yi Dou}, \bibinfo{person}{Yichong Xu}, \bibinfo{person}{Zhe Gan}, \bibinfo{person}{Jianfeng Wang}, \bibinfo{person}{Shuohang Wang}, \bibinfo{person}{Lijuan Wang}, \bibinfo{person}{Chenguang Zhu}, \bibinfo{person}{Pengchuan Zhang}, \bibinfo{person}{Lu Yuan}, \bibinfo{person}{Nanyun Peng}, {et~al\mbox{.}}} \bibinfo{year}{2022}\natexlab{}.
\newblock \showarticletitle{An empirical study of training end-to-end vision-and-language transformers}. In \bibinfo{booktitle}{\emph{Proceedings of the IEEE/CVF Conference on Computer Vision and Pattern Recognition}}. \bibinfo{pages}{18166--18176}.
\newblock


\bibitem[Gui et~al\mbox{.}(2024)]%
        {gui2024calibration}
\bibfield{author}{\bibinfo{person}{Xiaoqiang Gui}, \bibinfo{person}{Yueyao Cheng}, \bibinfo{person}{Xiang-Rong Sheng}, \bibinfo{person}{Yunfeng Zhao}, \bibinfo{person}{Guoxian Yu}, \bibinfo{person}{Shuguang Han}, \bibinfo{person}{Yuning Jiang}, \bibinfo{person}{Jian Xu}, {and} \bibinfo{person}{Bo Zheng}.} \bibinfo{year}{2024}\natexlab{}.
\newblock \showarticletitle{Calibration-compatible Listwise Distillation of Privileged Features for CTR Prediction}. In \bibinfo{booktitle}{\emph{Proceedings of the 17th ACM International Conference on Web Search and Data Mining}}. \bibinfo{pages}{247--256}.
\newblock


\bibitem[Han et~al\mbox{.}(2018)]%
        {han2018co}
\bibfield{author}{\bibinfo{person}{Bo Han}, \bibinfo{person}{Quanming Yao}, \bibinfo{person}{Xingrui Yu}, \bibinfo{person}{Gang Niu}, \bibinfo{person}{Miao Xu}, \bibinfo{person}{Weihua Hu}, \bibinfo{person}{Ivor Tsang}, {and} \bibinfo{person}{Masashi Sugiyama}.} \bibinfo{year}{2018}\natexlab{}.
\newblock \showarticletitle{Co-teaching: Robust training of deep neural networks with extremely noisy labels}.
\newblock \bibinfo{journal}{\emph{Advances in neural information processing systems}}  \bibinfo{volume}{31} (\bibinfo{year}{2018}), \bibinfo{pages}{8536–8546}.
\newblock


\bibitem[Hinton et~al\mbox{.}(2015)]%
        {hinton2015distilling}
\bibfield{author}{\bibinfo{person}{Geoffrey Hinton}, \bibinfo{person}{Oriol Vinyals}, {and} \bibinfo{person}{Jeff Dean}.} \bibinfo{year}{2015}\natexlab{}.
\newblock \showarticletitle{Distilling the knowledge in a neural network}.
\newblock \bibinfo{journal}{\emph{arXiv preprint arXiv:1503.02531}} (\bibinfo{year}{2015}).
\newblock


\bibitem[Li et~al\mbox{.}(2022)]%
        {li2022blip}
\bibfield{author}{\bibinfo{person}{Junnan Li}, \bibinfo{person}{Dongxu Li}, \bibinfo{person}{Caiming Xiong}, {and} \bibinfo{person}{Steven Hoi}.} \bibinfo{year}{2022}\natexlab{}.
\newblock \showarticletitle{Blip: Bootstrapping language-image pre-training for unified vision-language understanding and generation}. In \bibinfo{booktitle}{\emph{International conference on machine learning}}. PMLR, \bibinfo{pages}{12888--12900}.
\newblock


\bibitem[Li et~al\mbox{.}(2021)]%
        {li2021align}
\bibfield{author}{\bibinfo{person}{Junnan Li}, \bibinfo{person}{Ramprasaath Selvaraju}, \bibinfo{person}{Akhilesh Gotmare}, \bibinfo{person}{Shafiq Joty}, \bibinfo{person}{Caiming Xiong}, {and} \bibinfo{person}{Steven Chu~Hong Hoi}.} \bibinfo{year}{2021}\natexlab{}.
\newblock \showarticletitle{Align before fuse: Vision and language representation learning with momentum distillation}.
\newblock \bibinfo{journal}{\emph{Advances in neural information processing systems}}  \bibinfo{volume}{34} (\bibinfo{year}{2021}), \bibinfo{pages}{9694--9705}.
\newblock


\bibitem[Lopez-Paz et~al\mbox{.}(2015)]%
        {lopez2015unifying}
\bibfield{author}{\bibinfo{person}{David Lopez-Paz}, \bibinfo{person}{L{\'e}on Bottou}, \bibinfo{person}{Bernhard Sch{\"o}lkopf}, {and} \bibinfo{person}{Vladimir Vapnik}.} \bibinfo{year}{2015}\natexlab{}.
\newblock \showarticletitle{Unifying distillation and privileged information}.
\newblock \bibinfo{journal}{\emph{arXiv preprint arXiv:1511.03643}} (\bibinfo{year}{2015}).
\newblock


\bibitem[Ortiz-Jimenez et~al\mbox{.}(2023)]%
        {ortiz2023does}
\bibfield{author}{\bibinfo{person}{Guillermo Ortiz-Jimenez}, \bibinfo{person}{Mark Collier}, \bibinfo{person}{Anant Nawalgaria}, \bibinfo{person}{Alexander~Nicholas D’Amour}, \bibinfo{person}{Jesse Berent}, \bibinfo{person}{Rodolphe Jenatton}, {and} \bibinfo{person}{Efi Kokiopoulou}.} \bibinfo{year}{2023}\natexlab{}.
\newblock \showarticletitle{When does privileged information explain away label noise?}. In \bibinfo{booktitle}{\emph{International Conference on Machine Learning}}. PMLR, \bibinfo{pages}{26646--26669}.
\newblock


\bibitem[Page et~al\mbox{.}(2023)]%
        {page2023node}
\bibfield{author}{\bibinfo{person}{Pranav~S Page}, \bibinfo{person}{Anand~S Siyote}, \bibinfo{person}{Vivek~S Borkar}, {and} \bibinfo{person}{Gaurav~S Kasbekar}.} \bibinfo{year}{2023}\natexlab{}.
\newblock \showarticletitle{Node Cardinality Estimation in the Internet of Things Using Privileged Feature Distillation}.
\newblock \bibinfo{journal}{\emph{arXiv preprint arXiv:2310.18664}} (\bibinfo{year}{2023}).
\newblock


\bibitem[Qi et~al\mbox{.}(2023)]%
        {qi2023fakesv}
\bibfield{author}{\bibinfo{person}{Peng Qi}, \bibinfo{person}{Yuyan Bu}, \bibinfo{person}{Juan Cao}, \bibinfo{person}{Wei Ji}, \bibinfo{person}{Ruihao Shui}, \bibinfo{person}{Junbin Xiao}, \bibinfo{person}{Danding Wang}, {and} \bibinfo{person}{Tat-Seng Chua}.} \bibinfo{year}{2023}\natexlab{}.
\newblock \showarticletitle{Fakesv: A multimodal benchmark with rich social context for fake news detection on short video platforms}. In \bibinfo{booktitle}{\emph{Proceedings of the AAAI Conference on Artificial Intelligence}}, Vol.~\bibinfo{volume}{37}. \bibinfo{pages}{14444--14452}.
\newblock


\bibitem[Radford et~al\mbox{.}(2021)]%
        {radford2021learning}
\bibfield{author}{\bibinfo{person}{Alec Radford}, \bibinfo{person}{Jong~Wook Kim}, \bibinfo{person}{Chris Hallacy}, \bibinfo{person}{Aditya Ramesh}, \bibinfo{person}{Gabriel Goh}, \bibinfo{person}{Sandhini Agarwal}, \bibinfo{person}{Girish Sastry}, \bibinfo{person}{Amanda Askell}, \bibinfo{person}{Pamela Mishkin}, \bibinfo{person}{Jack Clark}, {et~al\mbox{.}}} \bibinfo{year}{2021}\natexlab{}.
\newblock \showarticletitle{Learning transferable visual models from natural language supervision}. In \bibinfo{booktitle}{\emph{International conference on machine learning}}. PMLR, \bibinfo{pages}{8748--8763}.
\newblock


\bibitem[Saito and Rehmsmeier(2015)]%
        {saito2015precision}
\bibfield{author}{\bibinfo{person}{Takaya Saito} {and} \bibinfo{person}{Marc Rehmsmeier}.} \bibinfo{year}{2015}\natexlab{}.
\newblock \showarticletitle{The precision-recall plot is more informative than the ROC plot when evaluating binary classifiers on imbalanced datasets}.
\newblock \bibinfo{journal}{\emph{PloS one}} \bibinfo{volume}{10}, \bibinfo{number}{3} (\bibinfo{year}{2015}), \bibinfo{pages}{e0118432}.
\newblock


\bibitem[Shu et~al\mbox{.}(2019)]%
        {shu2019meta}
\bibfield{author}{\bibinfo{person}{Jun Shu}, \bibinfo{person}{Qi Xie}, \bibinfo{person}{Lixuan Yi}, \bibinfo{person}{Qian Zhao}, \bibinfo{person}{Sanping Zhou}, \bibinfo{person}{Zongben Xu}, {and} \bibinfo{person}{Deyu Meng}.} \bibinfo{year}{2019}\natexlab{}.
\newblock \showarticletitle{Meta-weight-net: Learning an explicit mapping for sample weighting}.
\newblock \bibinfo{journal}{\emph{Advances in neural information processing systems}}  \bibinfo{volume}{32} (\bibinfo{year}{2019}), \bibinfo{numpages}{12}~pages.
\newblock


\bibitem[Wang et~al\mbox{.}(2021)]%
        {wang2021simvlm}
\bibfield{author}{\bibinfo{person}{Zirui Wang}, \bibinfo{person}{Jiahui Yu}, \bibinfo{person}{Adams~Wei Yu}, \bibinfo{person}{Zihang Dai}, \bibinfo{person}{Yulia Tsvetkov}, {and} \bibinfo{person}{Yuan Cao}.} \bibinfo{year}{2021}\natexlab{}.
\newblock \showarticletitle{SimVLM: Simple Visual Language Model Pretraining with Weak Supervision}. In \bibinfo{booktitle}{\emph{International Conference on Learning Representations}}.
\newblock


\bibitem[Wei et~al\mbox{.}(2020)]%
        {wei2020combating}
\bibfield{author}{\bibinfo{person}{Hongxin Wei}, \bibinfo{person}{Lei Feng}, \bibinfo{person}{Xiangyu Chen}, {and} \bibinfo{person}{Bo An}.} \bibinfo{year}{2020}\natexlab{}.
\newblock \showarticletitle{Combating noisy labels by agreement: A joint training method with co-regularization}. In \bibinfo{booktitle}{\emph{Proceedings of the IEEE/CVF conference on computer vision and pattern recognition}}. \bibinfo{pages}{13726--13735}.
\newblock


\bibitem[Xu et~al\mbox{.}(2020)]%
        {xu2020privileged}
\bibfield{author}{\bibinfo{person}{Chen Xu}, \bibinfo{person}{Quan Li}, \bibinfo{person}{Junfeng Ge}, \bibinfo{person}{Jinyang Gao}, \bibinfo{person}{Xiaoyong Yang}, \bibinfo{person}{Changhua Pei}, \bibinfo{person}{Fei Sun}, \bibinfo{person}{Jian Wu}, \bibinfo{person}{Hanxiao Sun}, {and} \bibinfo{person}{Wenwu Ou}.} \bibinfo{year}{2020}\natexlab{}.
\newblock \showarticletitle{Privileged features distillation at taobao recommendations}. In \bibinfo{booktitle}{\emph{Proceedings of the 26th ACM SIGKDD International Conference on Knowledge Discovery \& Data Mining}}. \bibinfo{pages}{2590--2598}.
\newblock


\bibitem[Yang et~al\mbox{.}(2022)]%
        {yang2022toward}
\bibfield{author}{\bibinfo{person}{Shuo Yang}, \bibinfo{person}{Sujay Sanghavi}, \bibinfo{person}{Holakou Rahmanian}, \bibinfo{person}{Jan Bakus}, {and} \bibinfo{person}{Vishwanathan SVN}.} \bibinfo{year}{2022}\natexlab{}.
\newblock \showarticletitle{Toward understanding privileged features distillation in learning-to-rank}.
\newblock \bibinfo{journal}{\emph{Advances in Neural Information Processing Systems}}  \bibinfo{volume}{35} (\bibinfo{year}{2022}), \bibinfo{pages}{26658--26670}.
\newblock


\bibitem[Zeng et~al\mbox{.}(2022)]%
        {zeng2022multi}
\bibfield{author}{\bibinfo{person}{Yan Zeng}, \bibinfo{person}{Xinsong Zhang}, {and} \bibinfo{person}{Hang Li}.} \bibinfo{year}{2022}\natexlab{}.
\newblock \showarticletitle{Multi-Grained Vision Language Pre-Training: Aligning Texts with Visual Concepts}. In \bibinfo{booktitle}{\emph{International Conference on Machine Learning}}. PMLR, \bibinfo{pages}{25994--26009}.
\newblock


\bibitem[Zhang et~al\mbox{.}(2022)]%
        {zhang2022confidence}
\bibfield{author}{\bibinfo{person}{Hailin Zhang}, \bibinfo{person}{Defang Chen}, {and} \bibinfo{person}{Can Wang}.} \bibinfo{year}{2022}\natexlab{}.
\newblock \showarticletitle{Confidence-aware multi-teacher knowledge distillation}. In \bibinfo{booktitle}{\emph{ICASSP 2022-2022 IEEE International Conference on Acoustics, Speech and Signal Processing (ICASSP)}}. IEEE, \bibinfo{pages}{4498--4502}.
\newblock


\bibitem[Zhang et~al\mbox{.}(2021)]%
        {zhang2021confidence}
\bibfield{author}{\bibinfo{person}{Songyuan Zhang}, \bibinfo{person}{Zhangjie Cao}, \bibinfo{person}{Dorsa Sadigh}, {and} \bibinfo{person}{Yanan Sui}.} \bibinfo{year}{2021}\natexlab{}.
\newblock \showarticletitle{Confidence-aware imitation learning from demonstrations with varying optimality}.
\newblock \bibinfo{journal}{\emph{Advances in Neural Information Processing Systems}}  \bibinfo{volume}{34} (\bibinfo{year}{2021}), \bibinfo{pages}{12340--12350}.
\newblock


\bibitem[Zhang et~al\mbox{.}(2020)]%
        {zhang2020prime}
\bibfield{author}{\bibinfo{person}{Youcai Zhang}, \bibinfo{person}{Zhonghao Lan}, \bibinfo{person}{Yuchen Dai}, \bibinfo{person}{Fangao Zeng}, \bibinfo{person}{Yan Bai}, \bibinfo{person}{Jie Chang}, {and} \bibinfo{person}{Yichen Wei}.} \bibinfo{year}{2020}\natexlab{}.
\newblock \showarticletitle{Prime-aware adaptive distillation}. In \bibinfo{booktitle}{\emph{Computer Vision--ECCV 2020: 16th European Conference, Glasgow, UK, August 23--28, 2020, Proceedings, Part XIX 16}}. Springer, \bibinfo{pages}{658--674}.
\newblock


\bibitem[Zhao et~al\mbox{.}(2023)]%
        {zhao2022exploring}
\bibfield{author}{\bibinfo{person}{Xiaoyan Zhao}, \bibinfo{person}{Min Yang}, \bibinfo{person}{Qiang Qu}, \bibinfo{person}{Ruifeng Xu}, {and} \bibinfo{person}{Jieke Li}.} \bibinfo{year}{2023}\natexlab{}.
\newblock \showarticletitle{Exploring privileged features for relation extraction with contrastive student-teacher learning}.
\newblock \bibinfo{journal}{\emph{IEEE Transactions on Knowledge and Data Engineering}} \bibinfo{volume}{35}, \bibinfo{number}{8} (\bibinfo{year}{2023}), \bibinfo{pages}{7953--7965}.
\newblock


\end{thebibliography}


\end{document}